\documentclass[journal]{IEEEtran}

\ifCLASSINFOpdf
\else
   \usepackage[dvips]{graphicx}
\fi
\usepackage{url}

\usepackage{graphicx}
\usepackage{amsmath,amssymb}
\usepackage{float}
\usepackage[caption = false, font=footnotesize]{subfig}
\usepackage{multirow}
\usepackage{xcolor}
\usepackage{balance}
\usepackage{makecell}
\usepackage{multicol} 

\begin{document}

\title{A Lightweight Dual-Mode Optimization for Generative Face Video Coding}

\author{Zihan Zhang,
        Shanzhi Yin, 
        Bolin Chen,~\IEEEmembership{Member,~IEEE}, 
        Ru-Ling Liao,
        Shiqi Wang,~\IEEEmembership{Senior Member,~IEEE},   
        Yan Ye,~\IEEEmembership{Senior Member,~IEEE}

\thanks{Zihan Zhang, Shanzhi Yin and Shiqi Wang are with the Department of Computer Science, City University of Hong Kong, Hong Kong (e-mail: zhzhang38-c@my.cityu.edu.hk, shanzhyin3-c@my.cityu.edu.hk and shiqwang@cityu.edu.hk).}
\thanks{Bolin Chen is with DAMO Academy, Alibaba Group, Hangzhou, China; Hupan Lab, Hangzhou, China; and Fudan University, Shanghai, China
(e-mail: chenbolin.chenboli@alibaba-inc.com).}
\thanks{Ru-Ling Liao and Yan Ye are with DAMO Academy, Alibaba Group, Sunnyvale, CA 94085 USA (e-mail: ruling.lrl@alibaba-inc.com and yan.ye@alibaba-inc.com).}
}

\markboth{Journal of \LaTeX\ Class Files, Vol. 14, No. 8, August 2015}
{Shell \MakeLowercase{\textit{et al.}}: Bare Demo of IEEEtran.cls for IEEE Journals}
\maketitle

\begin{abstract}

Generative Face Video Coding (GFVC) achieves superior rate-distortion performance by leveraging the strong inference capabilities of deep generative models. 
However, its practical deployment is hindered by large model parameters and high computational costs.
To address this, we propose a lightweight GFVC framework that introduces dual-mode optimization—combining architectural redesign and operational refinement—to reduce complexity whilst preserving reconstruction quality. 
Architecturally, we replace traditional $3 \times 3$ convolutions with slimmer and more efficient layers, reducing complexity without compromising feature expressiveness.
Operationally, we develop a two-stage adaptive channel pruning strategy: (1) soft pruning during training identifies redundant channels via learnable thresholds, and (2) hard pruning permanently eliminates these channels post-training using a derived mask. This dual-phase approach ensures both training stability and inference efficiency.
Experimental results demonstrate that the proposed lightweight dual-mode optimization for GFVC can achieve 90.4\% parameter reduction and 88.9\% computation saving compared to the baseline, whilst achieving superior performance compared to state-of-the-art video coding standard Versatile Video Coding~(VVC) in terms of perceptual-level quality metrics.
As such, the proposed method is expected to enable efficient GFVC deployment in resource-constrained environments such as mobile edge devices.

\end{abstract}


\IEEEpeerreviewmaketitle

\section{Introduction}
\IEEEPARstart{T}{he} increasing demand for video conferencing has driven the need for efficient transmission and compression of facial videos.
In recent years, traditional video compression standards, such as H.265/High Efficiency Video Coding (HEVC) ~\cite{sullivan2012overview} and H.266/Versatile Video Coding (VVC)~\cite{Bross2021vvc}, have been developed to improve compression efficiency while maintaining high visual quality.
Simultaneously, the emergence of generative models~\cite{chen2024generative}, especially Generative Adversarial Networks (GANs)~\cite{goodfellow2020generative}, has significantly advanced facial video compression techniques.
Generative Face Video Compression (GFVC)~\cite{chen2023generative,11002361,10109861,yin2024parametertranslator} employs compact representations at the encoder to model facial motion and deep generative models to reconstruct high-quality video at the decoder, achieving ultra-low bit-rate facial video transmission with vivid reconstructions~\cite{chen2024standardizing,chen2025generative}.
Despite its impressive rate-distortion (RD) performance, GFVC suffers from high model complexity, posing challenges for deployment on resource-limited devices.

\begin{figure}[t]
\centering
\includegraphics[width=0.5\textwidth]{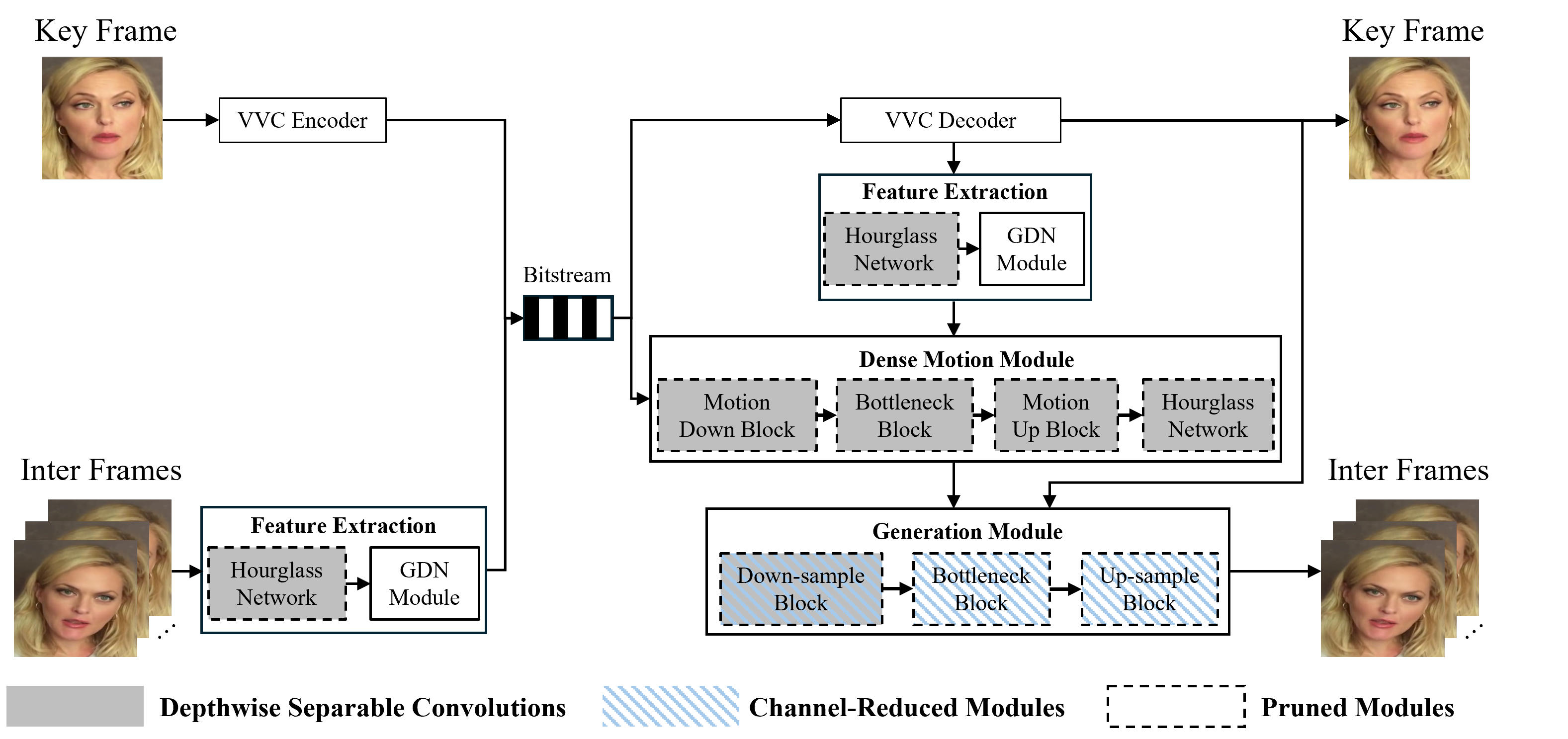}
\caption{Overview of the proposed lightweight GFVC framework. The lightweight GFVC system integrates architectural and operational optimization. Architectural optimization includes replacing standard convolutions with depthwise separable convolutions (gray) and reducing channels in the generation module (blue-striped). Operational optimization introduces pruning modules (dashed boxes) that identify and eliminate redundant channels, contributing to model efficiency.}
\label{fig:gfvc}
\vspace{-1.5em}
\end{figure}

\begin{table*}[t]
\centering
\vspace{-2em}
\scriptsize
\renewcommand{\arraystretch}{1.5}
\caption{Distribution of Parameters(M) across CFTE modules}
\label{tab:gfvc_modules}
\setlength{\tabcolsep}{3pt}
\begin{tabular}{c|cc|ccc|ccc}
\hline
\textbf{Module} 
& \multicolumn{2}{c|}{\textbf{Feature Extraction}} 
& \multicolumn{3}{c|}{\textbf{Dense Motion Module}} 
& \multicolumn{2}{c}{\textbf{Generation Module}} \\
\cline{1-8}
\textbf{Network Structure} 
& \textbf{Hourglass Network} & \textbf{GDN Module} 
& \textbf{Motion Down/Up Block} & \textbf{Bottleneck Block} & \textbf{Hourglass Network} 
& \textbf{Down/Up-sample Block} & \textbf{Bottleneck Block} \\
\hline
Original      
& 13.21M & 0.92M
& 7.82M & 18.89M & 9.31M & 0.36M & 6.50M \\
Proposed w/o pruning  
& 0.70M &  0.92M
& 0.90M & 1.32M & 1.07M & 0.10M & 1.63M \\
Proposed      & 0.60M  & 0.92M & 0.68M & 1.32M & 0.61M &  0.10M & 0.54M \\
\hline
\end{tabular}
\vspace{-1.8em}
\end{table*}

To address computational challenges of deep learning models, lightweight  techniques have been explored to maintain performance with reduced complexity.
Notable examples include MobileNet~\cite{howard2017mobilenets,sandler2018mobilenetv2,howard2019searching}, EfficientViT~\cite{cai2023efficientvit} and LDNN-PFD~\cite{liu2024lightweight}, which use efficient convolution operations to reduce parameter amount and complexity. 
For instance, Light-weighted Temporal Evolution Inference (LW-TEI)~\cite{zhang2024light} employs inverted residual structures to reduce parameters in GFVC models, followed by feature-level knowledge distillation to enhance performance. However, this approach achieves limited parameter reduction and degrades performance in complex GFVC scenarios. Another common approach is model pruning, which eliminates redundant weights, channels, or layers to reduce memory and computation cost. However, most traditional pruning techniques~\cite{liu2019channel,li2024pruningbench, wu2025pr, lu2025novel} require manually specified configurations, such as fixed pruning ratios and thresholds, which may lack flexibility and generalizability.

Although some pruning methods avoid manual configurations by learning pruning ratios or thresholds automatically~\cite{saxena2024rg,ma2024global,wu2025quantcache}, their application in the highly complex domain of facial video compression remains limited. 
This is largely due to the multifaceted nature of GFVC models, which contain various modules and complete diverse tasks in an end-to-end pipeline, such as feature extraction, motion estimation and image generation, which span both low-level vision and high-level semantics.
Furthermore, GFVC models often involve intricate designs with residual connections, skip layers, and cross-modules dependencies, which can cause dimensional mismatches if pruning is applied indiscriminately.

In this paper, we propose a lightweight generative face video compression framework that reduces model complexity while maintaining reconstruction quality. The framework is built on Compact Feature Temporal Evolution (CFTE)~\cite{chen2022beyond}, one of the representative baselines in GFVC.
Specifically, CFTE models the temporal evolution between frames using a compact $4 \times 4$ feature matrix for each inter frame, which is extracted and compressed at the encoder side and used to guide frame synthesis via dense motion and occlusion maps at the decoder side. This design enables ultra-low bitrate compression while preserving high perceptual quality.
For architectural optimization, we replace standard $3 \times 3$ convolutions in CFTE’s parameter-heavy structures with slimmer depthwise separable convolutions, significantly reducing the number of parameters and computational cost. For operational optimization, we introduce a two-stage adaptive channel pruning strategy. In the first stage, soft pruning is applied during training using learnable thresholds to identify redundant channels. In the second stage, hard pruning removes these channels based on the final learned masks.
The major contribution of this work can be summarized as follows:
\begin{itemize}
    \item We propose a lightweight generative face video coding framework that effectively reduces model parameters and computational complexity through dual-mode optimization combining architectural redesign and operational refinement.
    \item We analyze the parameter and computation distribution across CFTE modules, and introduce targeted architectural optimizations using depthwise separable convolutions and channel slimming to reduce redundancy in modules with high computational cost.
    \item We develop a structured pruning approach that embeds a differentiable binary mask mechanism into BatchNorm layers. As such, it enables adaptive and automated channel pruning during training, followed by hard pruning and post-training, yielding efficient models without compromising reconstruction quality.
\end{itemize}

\begin{figure}[t]
    \centering
    \includegraphics[width=0.48\textwidth]{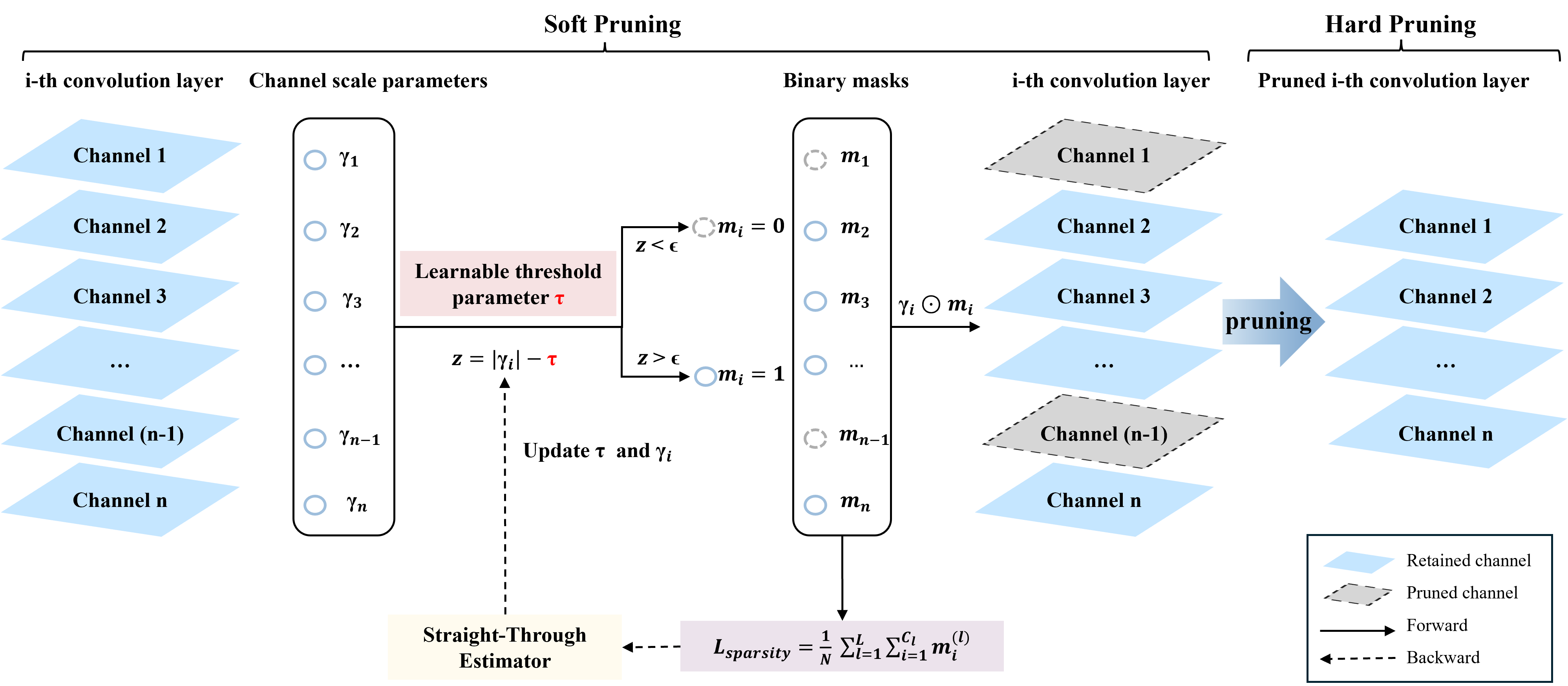}
    \caption{Adaptive channel pruning for operational optimization. The soft pruning strategy is adopted by learning channel scaling factors and binary masks with a learnable threshold \(\tau\). A sparsity loss \(L_{\text{sparsity}}\) guides the pruning mask learning via a straight-through estimator, followed by hard pruning to remove unimportant channels from the convolutional layers, thereby improving efficiency without significant accuracy loss.} 
    \label{fig:pruning}
    \vspace{-1.8em}
\end{figure}

\section{Proposed Method}
In this section, we provide an overview of the proposed lightweight optimization strategy for generative face video coding. As illustrated in Fig.~\ref{fig:gfvc}, our approach consists of two key components: architectural optimization using depthwise separable convolutions and channel reduction, and operational optimization via adaptive channel pruning.
\subsection{Architectural Optimization with Efficient Convolutions and Channel Reduction}
\label{sec:Architectural Optimization}

To better understand where the computational burden lies, we analyze the parameter and computation distribution across CFTE modules. As shown in Table~\ref{tab:gfvc_modules}, most of the complexity is concentrated in the feature extraction and dense motion modules, particularly in the hourglass networks and bottleneck blocks.
Motivated by these observations, we explore lightweight architectural modifications to alleviate redundant computation while preserving performance.
Following previous work~\cite{JVET-AH0109}, which introduced lightweight strategies for CFTE, we adopt architectural optimization including depth-wise structures and channel reduction. Specifically, as illustrated in Fig.~\ref{fig:gfvc}, we apply depthwise separable convolutions to selected  $3 \times 3$ convolutions within the hourglass networks of both the feature extraction and dense motion modules. This yields a computational reduction by a factor of approximately $1/N + 1/D_K^2$, where $D_K$ is the kernel size, and $N$ is the number of output channels. The same replacement is also applied to the motion down/up and bottleneck blocks in the dense motion module. Additionally, we reduce the number of channels in the generation module to further alleviate computational costs.

Compared to the original CFTE model, our architectural optimization achieves substantial parameter reductions in high-complexity components, as summarized in Table~\ref{tab:gfvc_modules}. Specifically, the hourglass networks in the feature extraction and dense motion modules achieve parameter reductions of 94.6\% and 88.5\%, respectively, through the application of depthwise separable convolutions. Moreover, the number of parameters in the bottleneck blocks is significantly compressed—from 18.89M to 1.32M in the dense motion module via depthwise separable convolutions. And through channel reduction, the number of parameters in the bottleneck block in generation module is compressed from 6.50M to 1.63M. 
These results demonstrate that most of the complexity lies in the hourglass structures and bottleneck blocks, and our architectural optimization effectively targets these high-cost components.

\subsection{Operational Optimization with Adaptive Channel Pruning}
\label{sec:Operational Optimization}

Traditional network pruning methods typically rely on manually predefined pruning ratios or thresholds, which require extensive empirical tuning and may result in suboptimal pruning configurations. To address this limitation, we propose an adaptive channel pruning framework based on learnable thresholds, as illustrated in Fig.~\ref{fig:pruning}. This approach introduces a differentiable binary mask in BatchNorm layers to achieve structured pruning optimization.

Specifically, we design a modified BatchNorm layer, termed MaskedBatchNorm2d, which integrates a learnable threshold parameter to adaptively modulate the BatchNorm scale parameter, thereby controlling the sparsity of each channel. 
Inspired by Network Slimming~\cite{liu2017learning}—which treats the BatchNorm scale parameter as an indicator of channel importance—our method improves upon it by eliminating the need for manually setting global pruning ratios or thresholds. The threshold is jointly optimized with model parameters, allowing channel selection to be learned in a data-driven manner.
Therefore, we construct a channel-wise binary mask $m_i \in \{0,1\}$ based on the scale parameters $\gamma_i$ and the learned threshold \(\tau\):
\begin{equation}
z_i = |\gamma_i| - \tau, \quad m_i = \mathbb{I}(z_i \geq 0)
\label{eq:mask}
\end{equation}
where \(\mathbb{I}(\cdot)\) denotes the indicator function, and \(\tau\) is initialized to 0.1 in our experiments. The mask effectively zeros out channels where the scale parameter magnitude is below the threshold, enabling channel-wise pruning. 

Since the binary mask $m_i$ involves a non-differentiable indicator function, direct gradient computation is not feasible. To address this, we utilize the Straight-Through Estimator (STE)~\cite{bengio2013estimating} to approximate gradients during back-propagation. 
Inspired by the theoretical analysis in~\cite{yin2019understanding}, which shows that coarse gradients in STE correlate more positively with the true gradient near decision boundaries, we adopt a clipped variant of STE. 
Specifically, we allow gradients to propagate only within a small \(\epsilon\)-neighborhood of the threshold boundary to improve convergence stability and reduce estimation bias.
Within this band, we set \(\partial m_i / \partial z_i \approx 1\) and compute the following gradients,
the gradient approximations for the mask with respect to $\gamma_i$ and $\tau$ are given by:
\begin{equation}
\frac{\partial m_i}{\partial \gamma_i} \approx \begin{cases}
\text{sign}(\gamma_i), & \text{if } |z_i| \leq \epsilon \\
0, & \text{otherwise}
\end{cases}
\label{eq:grad_gamma}
\end{equation}

\begin{equation}
\frac{\partial m_i}{\partial \tau} \approx \begin{cases}
-1, & \text{if } |z_i| \leq \epsilon \\
0, & \text{otherwise}
\end{cases}
\label{eq:grad_tau}
\end{equation}
where \(\epsilon\) is a small constant that controls the width of the gradient propagation window. In our experiments, we set \(\epsilon = 1.0\) to balance gradient stability and learning efficiency.
The forward operation for the masked BatchNorm scale parameter $\gamma_i$ is defined as:
\begin{equation}
\hat{\gamma}_i = \gamma_i \odot m_i
\label{eq:x-hat}
\end{equation}
where $  \odot  $ denotes element-wise multiplication. This formulation ensures that channels with $m_i = 0$ are effectively deactivated by setting their corresponding scale to zero during forward propagation.

To encourage the model to learn a sparse channel configuration, we define the sparsity loss as:
\begin{equation}
\mathcal{L}_{\text{sparsity}} = \frac{1}{N} \sum_{l=1}^{L} \sum_{i=1}^{C_l} m_i^{(l)}
\label{eq:sparsity}
\end{equation}
where $N$ denotes the total number of channels across all layers, $L$ is the number of layers subject to pruning, $C_l$ is the number of channels in the $l$-th layer.
The overall training objective is defined as : 
\begin{equation}
\mathcal{L}_{\text{total}} = \mathcal{L}_{\text{task}} + \lambda_{\text{sparse}} \mathcal{L}_{\text{sparsity}}
\label{eq:L-total}
\end{equation}
where $\mathcal{L}_{\text{task}}$ denotes the primary loss function used for training the CFTE model, and $\lambda_{\text{sparse}}$ is a regularization coefficient that balances the sparsity constraint.
To prevent premature pruning and ensure stable convergence, we adopt a progressive training strategy, where $\lambda_{\text{sparse}} = 0$ in the initial phase, with the sparsity constraint gradually introduced in subsequent stages.

Moreover, our pruning process is designed as two stages. In the first stage, \textit{soft pruning} is performed during training: the network architecture remains unchanged, but redundant channels are dynamically suppressed through binary masks with STE-based gradient approximation in the proposed MaskedBatchNorm2d. During this phase, both the scale parameter $\gamma$ and threshold $\tau$ are jointly optimized—$\gamma$ is influenced by task-specific and sparsity losses, while $\tau$ is updated via the STE to gradient flow through the non-differentiable masking operation.
In the second stage, \textit{hard pruning} is performed after training by removing channels whose final mask value is zero. However, to preserve architectural integrity, channels that are structurally required by subsequent layers are retained, even if their mask values are zero. This results in a compact and efficient network architecture, followed by a fine-tuning step to refine the remaining channels and recover potential performance degradation introduced by pruning.
This approach removes the need for manual pruning ratio specification and allows the network to autonomously learn an optimal sparse structure based on data-driven criteria, while maintaining training stability through differentiability.

\begin{table}[t]
  \centering
  \vspace{-2em}
  \setlength{\tabcolsep}{1pt}
  \renewcommand{\arraystretch}{1.25}
  \scriptsize
  \caption{Comparison of model complexity and compression efficiency (vs. VVC anchor) for CFTE variants using different lightweight strategies}
  \label{tab:BD-rate saving}
  \begin{tabular}{c|cc|cc}
    \hline
    \textbf{Method}  & \textbf{Params (M)} & \textbf{KMACs/Pixel} & \textbf{Rate-DISTS} & \textbf{Rate-LPIPS} \\
    \hline
    Original & 58.03 & 852.00 & -56.68\% & -52.41\% \\
    LW-TEI & 36.85 & 381.93 & -57.19\% & -50.08\% \\
    JVET-AH0109 & 28.33 & 209.31 &  -55.54\% & -47.89\% \\ 
    Proposed (CFTE-Lite) w/o pruning & 7.65 & 174.24 & -56.76\% & -48.45\% \\ 
    Proposed (CFTE-Lite)  & 5.58 & 94.57 & -58.04\% & -51.87\% \\ 
    \hline
  \end{tabular}
  \vspace{-1.8em}
\end{table}

\begin{figure}[t]
\centering
\vspace{-1.5em}
\subfloat[ClassA Sequence 01 at 5kps]{
    \includegraphics[width=0.8\linewidth]{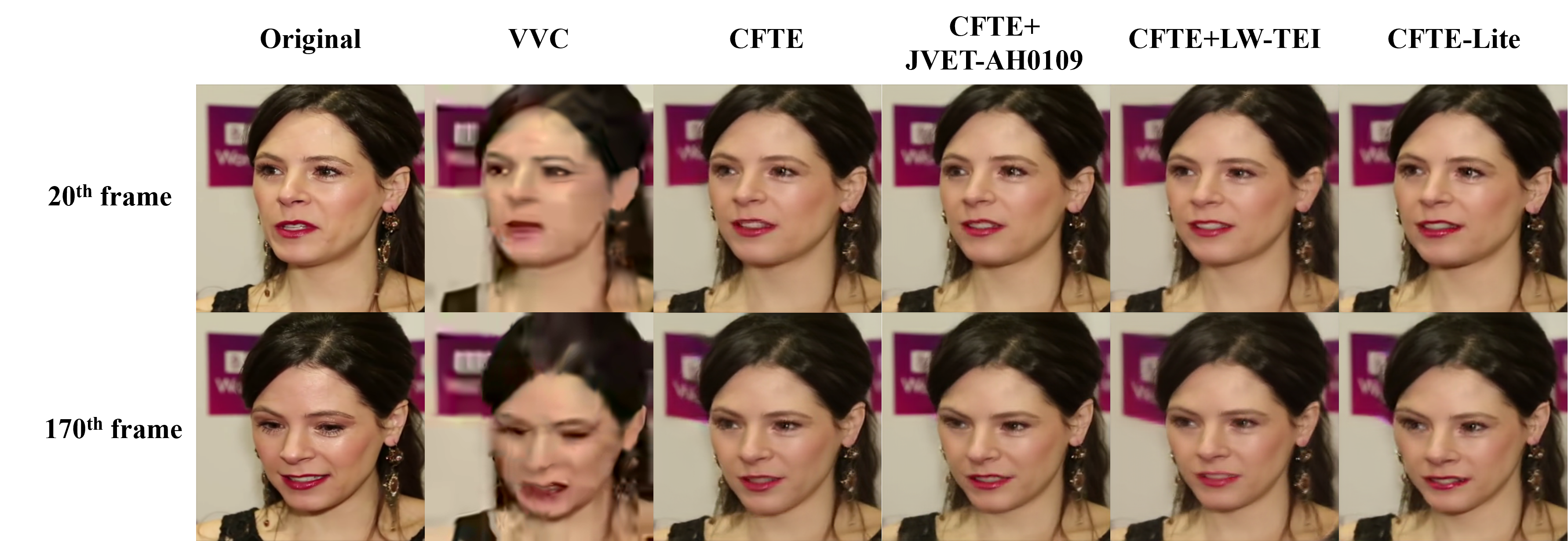}
    \label{fig:sub1}
}
\vspace{0.2em} 
\subfloat[ClassB Sequence 05 at 6kps]{
    \includegraphics[width=0.8\linewidth]{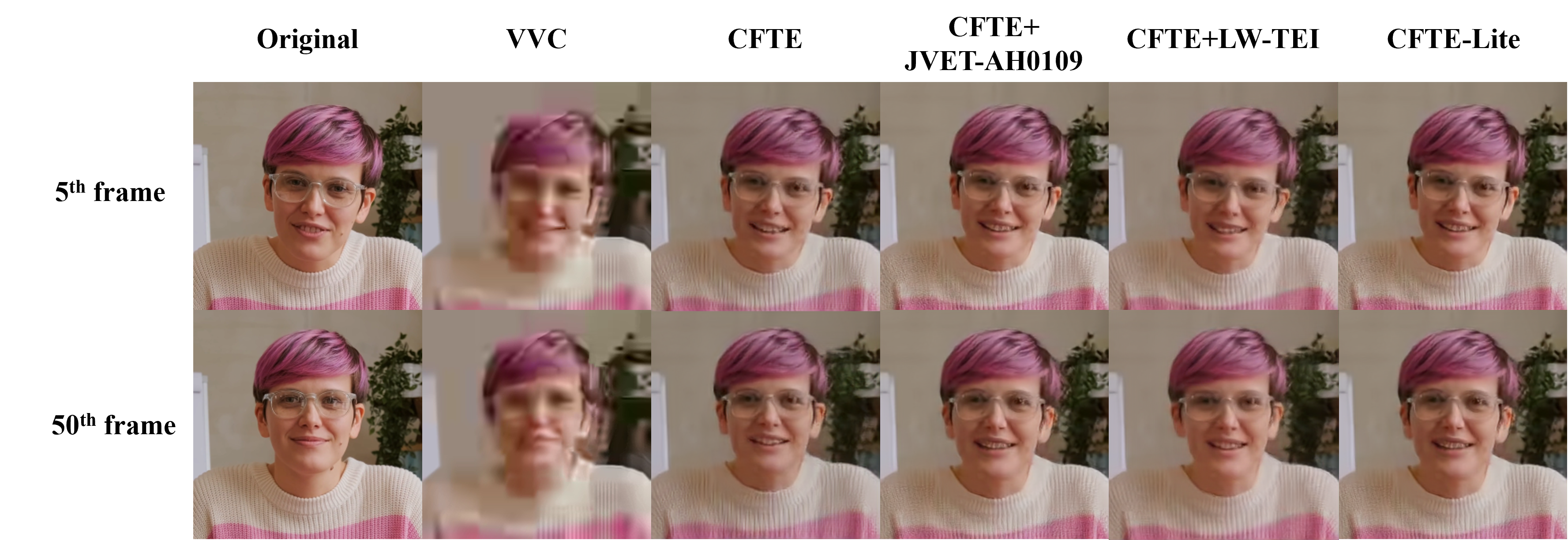}
    \label{fig:sub2}
}
\caption{Visual quality comparisons among VVC, CFTE and CFTE variants at similar bit rates for each sequence.} 
\label{fig:subjective}
\vspace{-2em}
\end{figure}

\begin{figure}[t]
\centering
\subfloat[Rate-DISTS]{\includegraphics[width=0.25\textwidth]{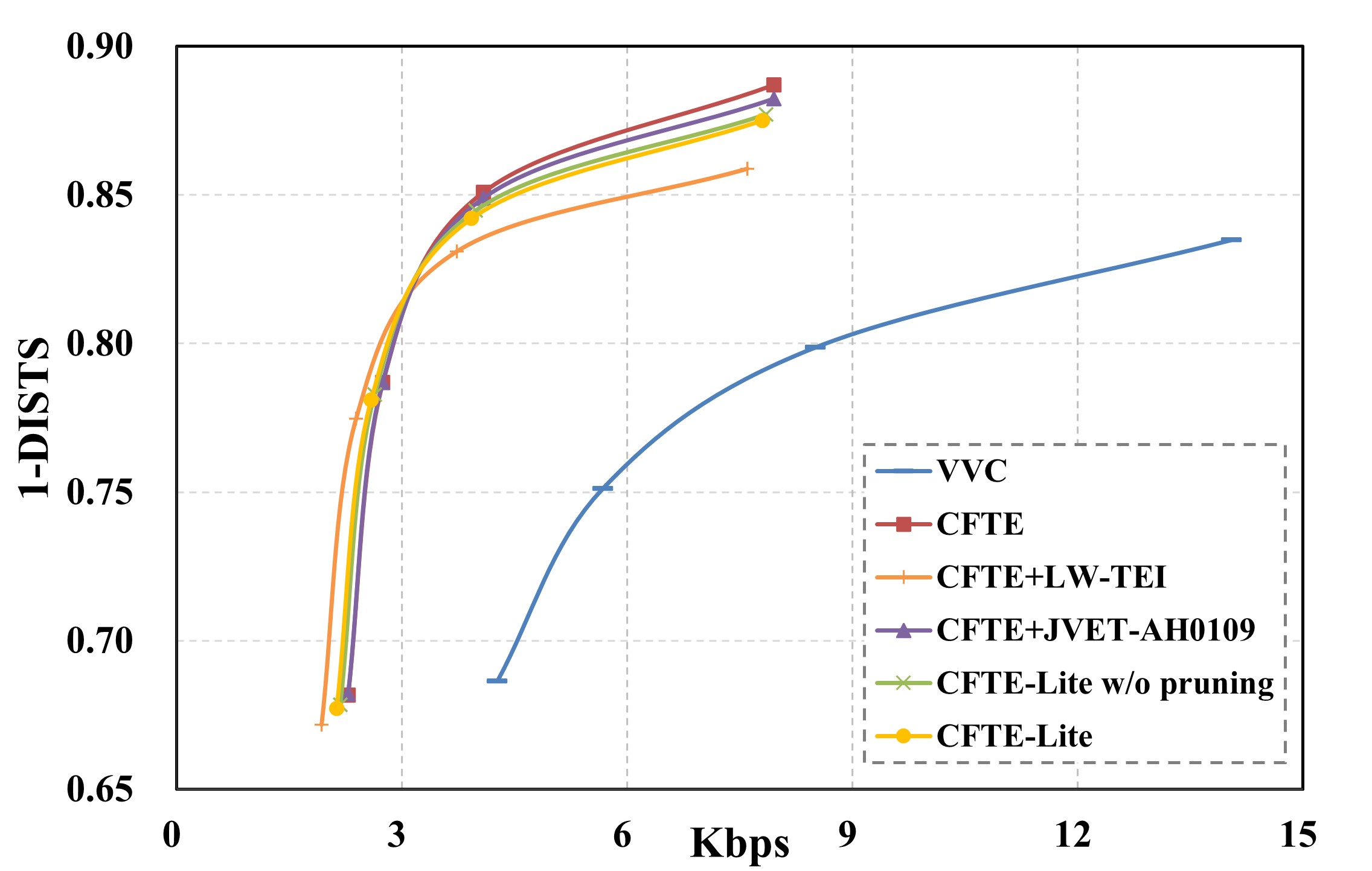}}
\subfloat[Rate-LPIPS]{\includegraphics[width=0.25\textwidth]{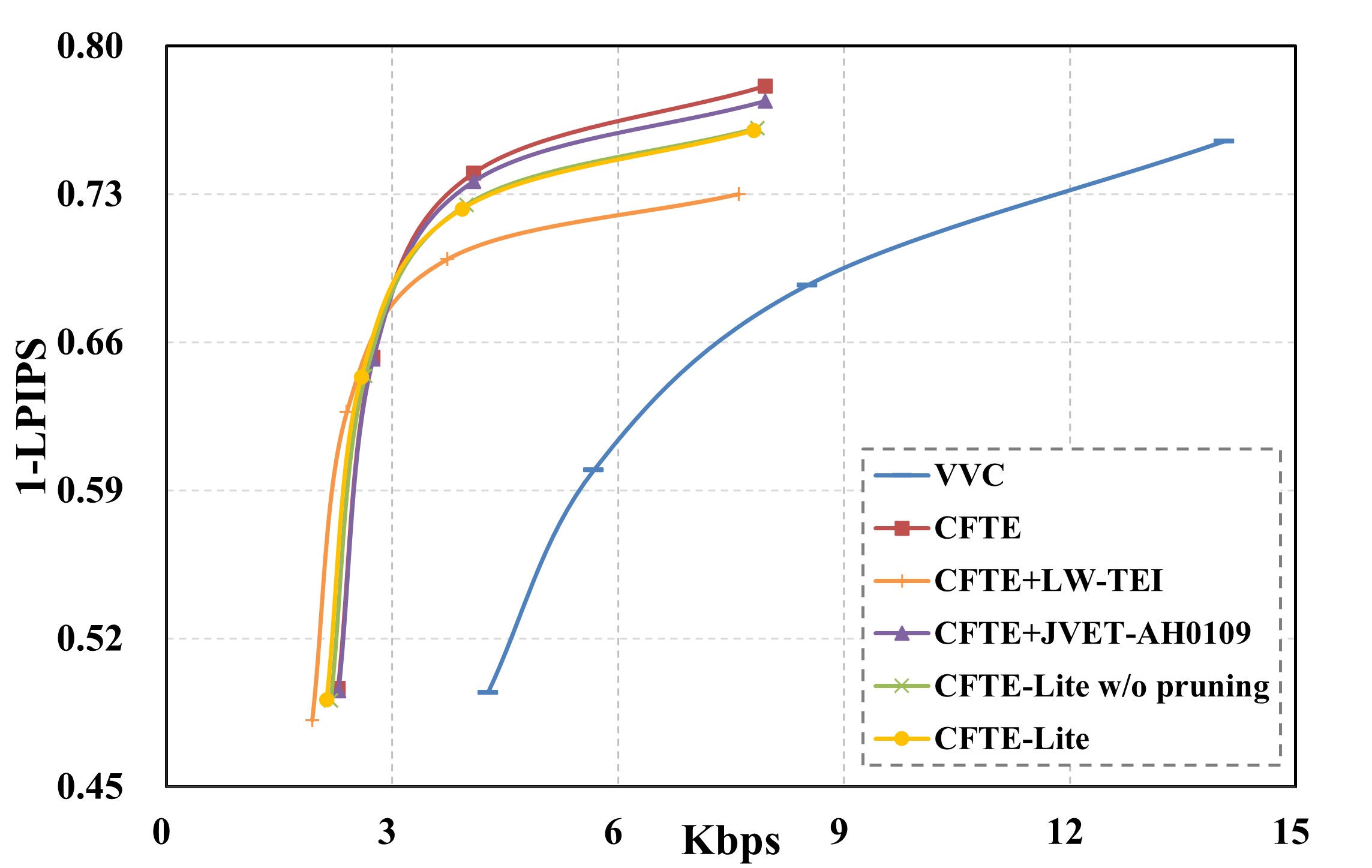}}
\caption{Rate-Distortion performance comparisons of VVC, CFTE and CFTE variants in terms of DISTS and LPIPS.} 
\label{fig:RD}
\vspace{-1.5em}
\end{figure}

\section{Experiments}
\subsection{Experimental Settings}
\subsubsection{Implementation Details}
The proposed lightweight CFTE model (CFTE-Lite) is trained on the VoxCeleb2~\cite{chung2018voxceleb2} and CelebVHQ~\cite{zhu2022celebvhq} datasets, using images with a resolution of $256 \times 256$. 
We implement the proposed model with the PyTorch framework and use NVIDIA TESLA A100 GPUs for training and fine-tuning. During training, the Adam optimizer is employed with parameters \(\beta_1 = 0.5\) and \(\beta_2 = 0.999\), and a learning rate of \(0.0002\) to ensure convergence. The sparsity coefficient $\lambda_{\text{sparse}}$ is initially set to 0 for the first 40 epochs and then increased to 0.1.
The model is trained for 100 epochs in total, followed by an additional 50 epochs of fine-tuning after hard pruning.
For performance evaluation, we follow the GFVC standardization activities in Joint Video Experts Team (JVET) of ISO/IEC JTC 1/SC 29 and ITU-T SG21 and adopt the test sequences specified in JVET-AG2035~\cite{JVET-AG2035}.
Specifically, the test set consists of 33 sequences at $256 \times 256$ resolution: Class A contains 15 face-centric sequences, each with 250 frames, and Class B includes 18 head-and-shoulder sequences, each with 125 frames.
\subsubsection{Compared Methods}
To demonstrate the effectiveness of the proposed CFTE-Lite, we compare its performance with the VVC, the original CFTE model, the CFTE model optimized with the lightweight approach proposed in LW-TEI~\cite{zhang2024light} and JVET-AH0109~\cite{JVET-AH0109}.
For VVC, we employ the VTM 22.2 reference software in Low-Delay Bidirectional mode with Quantization Parameters (QPs) set to 37, 42, 47, and 52. For CFTE and its variants, key-reference frames are encoded using VTM 22.2 with QPs of 22, 32, 42, and 52, and the subsequent inter frames are compressed using the GFVC solution.

\subsubsection{Evaluation Measures}
We employ two perceptual-level metrics to assess the quality of the generated videos: Deep Image Structure and Texture Similarity (DISTS)~\cite{ding2020image} and Learned Perceptual Image Patch Similarity (LPIPS)~\cite{zhang2018unreasonable}. Bjøntegaard Delta Rate (BD-rate)~\cite{Bjntegaard2001CalculationOA} and RD curve are adopted to quantify the overall compression performances of different models.
To present the RD curves in increasing order and calculate the savings of the BD-rate, we use ``1-DISTS'', ``1-LPIPS'' as the y-axis of the graphs.

\subsection{Comparisons}
\subsubsection{Model Complexity}

Table \ref{tab:BD-rate saving} compares the parameters and KMACs/pixel between the original CFTE, the LW-TEI CFTE variant, the JVET-AH0109 CFTE variant, and CFTE-Lite. The proposed CFTE-Lite requires only 5.58M parameters and 94.57 KMACs/pixel. This corresponds to 90.4\% parameters reduction and 88.9\% computation reduction compared to the original CFTE model, demonstrating the effectiveness of our optimization.

\subsubsection{Rate-Distortion Performance}
Table~\ref{tab:BD-rate saving} shows the BD-rate savings of the original CFTE, the LW-TEI CFTE variant, the JVET-AH0109 CFTE variant, and the proposed CFTE-Lite compared to the VVC anchor.  
CFTE-Lite achieves BD-rate savings of 58.04\% and 51.87\% for Rate-DISTS and Rate-LPIPS, respectively, outperforming both the LW-TEI CFTE variant and the JVET-AH0109 variant.
Figure~\ref{fig:RD} illustrates the RD performance in terms of DISTS and LPIPS for VVC, CFTE, the LW-TEI CFTE variant, the JVET-AH0109 CFTE variant, and the proposed CFTE-Lite. It can be observed that CFTE-Lite significantly outperforms VVC and the LW-TEI CFTE variant while exhibiting only slight performance degradation compared to CFTE, within an acceptable range. This demonstrates that CFTE-Lite effectively balances RD performance and model complexity.

\subsubsection{Subjective Quality}

As shown in Figure~\ref{fig:subjective}, we visualize two sequences under ultra-low bitrates. VVC results suffer from strong blocking artifacts, while the proposed CFTE-Lite still reconstructs coherent and realistic facial structures with a significantly reduced model size. In particular, the proposed CFTE-Lite reconstructions show negligible additional distortion compared to the original CFTE reconstructions.

\subsection{Ablation Studies}
To further validate the effectiveness of the proposed adaptive channel pruning method, we conduct an ablation study comparing the proposed CFTE-Lite with and without the pruning operation. 
It can be seen from Table~\ref{tab:BD-rate saving} that the pruning process reduces the parameter amount from 7.65M to 5.58M and the computational cost from 174.24 to 94.57 KMACs/pixel.
Fig.~\ref{fig:RD} shows the RD performance of CFTE-Lite without pruning and CFTE-Lite in terms of DISTS and LPIPS. It can be observed that the pruned model achieves nearly identical RD performance compared to the unpruned variant. This result demonstrates that the proposed adaptive channel pruning effectively removes redundant channels while preserving critical features for high-quality facial video reconstruction.

\section{Conclusion}

This paper proposes a lightweight generative face video compression framework that reduces model complexity through architectural and operational optimization. By integrating efficient depthwise convolutional layers and a two-stage adaptive pruning method, our approach achieves a 90.4\% reduction in parameters and 88.9\% savings in KMACs/pixel compared to the baseline CFTE model, while maintaining promising RD performance. These results demonstrate the potential of lightweight GFVC for future deployment on resource-constrained devices.

\begin{multicols}{2}
    \columnbreak
\end{multicols}

\balance
\bibliographystyle{IEEEtran}
\bibliography{main}

\end{document}